\documentclass{article}
\usepackage{spconf,amsmath,graphicx}
\usepackage[pagebackref,breaklinks,colorlinks]{hyperref}
\usepackage{xcolor}
\usepackage{booktabs}

\title{Integrating Audio Narrations to Strengthen Domain Generalization in Multimodal First-Person Action Recognition}
\name{
Cagri Gungor and Adriana Kovashka
}
\address{University of Pittsburgh}

\begin{document}
%
\maketitle
\begin{abstract}
First-person activity recognition is rapidly growing due to the widespread use of wearable cameras but faces challenges from domain shifts across different environments, such as varying objects or background scenes. We propose a multimodal framework that improves domain generalization by integrating motion, audio, and appearance features. Key contributions include analyzing the resilience of audio and motion features to domain shifts, using audio narrations for enhanced audio-text alignment, and applying consistency ratings between audio and visual narrations to optimize the impact of audio in recognition during training. Our approach achieves state-of-the-art performance on the ARGO1M dataset, effectively generalizing across unseen scenarios and locations.
\end{abstract}
\section{Introduction}
\label{sec:intro}

With the growing prevalence of wearable technology and first-person cameras, first-person activity recognition has emerged as a crucial area of research \cite{kazakos2019epic, gong2023mmg, zhao2023learning,wang2021interactive, plizzari2022e2}. This field is vital for real-world egocentric vision applications, from human-robot interaction to personalized assistance. However, a significant challenge in developing robust action recognition models is the issue of domain shifts—variations in environmental contexts, objects, and activities that can drastically affect the performance of these models. Some works \cite{li2017deeper,muandet2013domain,piratla2020efficient,zhang2017mixup,plizzari2023can} rely solely on visual features to generalize across different domains. In this paper, we explore the potential of multimodality, specifically the integration of motion and audio with appearance, to enhance domain generalization in first-person action recognition tasks. 

\begin{figure}[t]
    \centering
    \includegraphics[width=\linewidth]{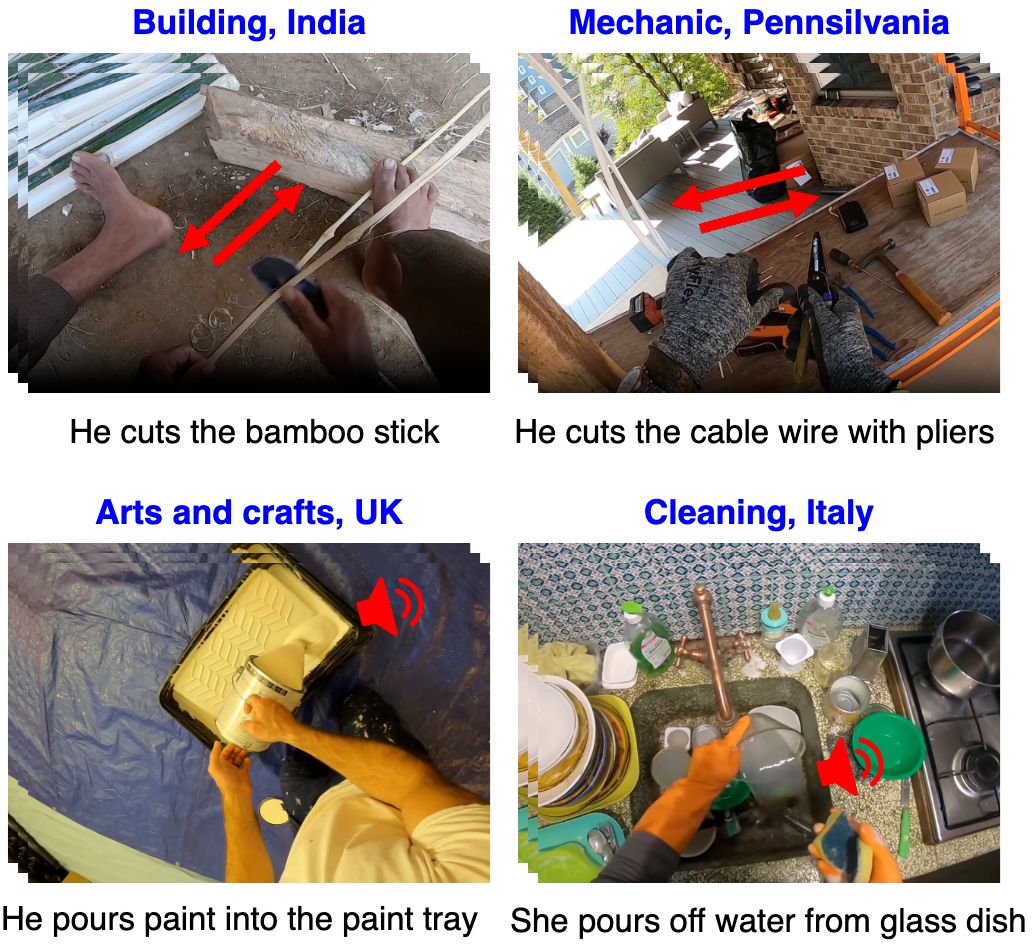}
    \caption{Illustration of motion and audio resilience to domain shifts compared to appearance. While the motion of `cutting' (first row) and audio of `pouring' (second row) remain similar across different scenario-location domains, the appearance varies significantly with different objects and backgrounds. }
 
    \label{fig:concept}
\end{figure}

Our intuition is that domain shifts arise heavily from variations in the spatial semantics of videos, such as changes in the type of objects or the background, which we refer to as the appearance modality. These variations can lead to a drop in model performance when applied to new, unseen data. However, the motion and audio modalities capture temporal dynamics that remain more consistent across different domains.

\begin{figure*}[t]
    \centering
    \includegraphics[width=0.9\linewidth]{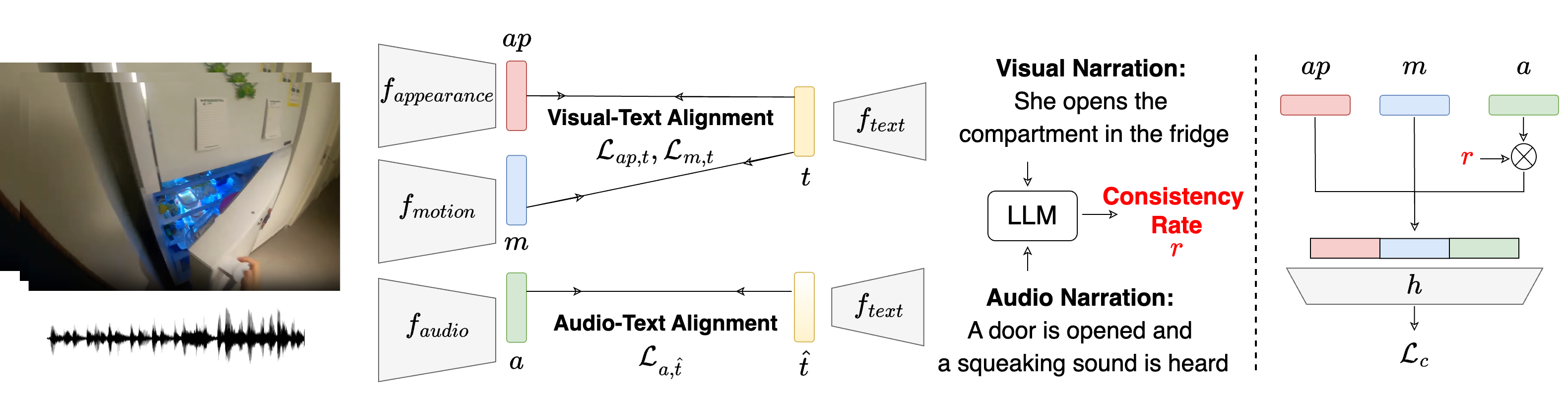}
    \vspace{-0.5cm}
    \caption{The proposed framework extracts appearance $ap_i$, motion $m_i$, and audio $a_i$ embeddings using trained encoders $f$. Visual-text and audio-text alignments are performed independently to enhance the robustness of action representations. Consistency rating $r_i$, calculated offline using a LLM \cite{touvron2023llama}, is then multiplied by audio embedding, optimizing the influence of audio in multimodal prediction. Note that narrations and consistency rate are only utilized during training to improve representation learning. During inference, embeddings are directly fused before prediction.  }
 
    \label{fig:main}
\end{figure*}

For example, while objects being `cut' in different domains, such as bamboo stick with knife (in building in India scenario) or cable wire with pliers (in the mechanic in Pennsylvania scenario) 
are visually different, the motion involved in cutting, characterized by the repetitive back-and-forth movement, remains consistent (Fig.~\ref{fig:concept} top). Similarly, `mixing' motions are consistent whether stirring ingredients or mixing cement. Audio also provides strong temporal cues that are less affected by visual changes. The sound of `pouring', whether paint into a tray or water from glass, remains consistent (Fig.~\ref{fig:concept} bottom). Similarly, the sound of `rolling', whether dough in a kitchen or clay in a studio, stays the same. Despite variations in visual context, the consistent motion patterns and audio cues are more reliable indicators of the action and are more robust to domain shifts. Our experimental analysis validate our intuition by showcasing lower performance drop of motion (25.8\%) and audio (32.7\%) comparing to appearance (54.8\%) when applied to unseen domains. 


In addition to leveraging motion and audio, our research also incorporates the text modality through narrations. Aligning \emph{visual} features with corresponding textual descriptions (in this case, narrations of actions) has been shown to enhance feature representation \cite{radford2021learning,
plizzari2023can, min2022grounding}. 
However, existing narrations are primarily based on visual content (in \cite{plizzari2023can}'s dataset, many videos do not contain audio, yet all contain narrations). 
Thus aligning \emph{audio} features with them
can introduce noise due to inconsistencies, as the actions depicted in the video may not always produce corresponding audio cues. 
To address this, we employ an audio captioner \cite{deshmukh2023pengi} to generate audio-specific narrations, enabling us to align appearance and motion with visual-based narrations, and audio with its respective narrations. Additionally, we calculate consistency ratings between visual and audio narrations using an LLM \cite{touvron2023llama}. This consistency indicates how closely the audio and video express the same action. We apply attention
weighing (Fig. 2) where lower consistency reduces the influence of audio in prediction. These approaches improve representation learning by reducing the noise and enhancing generalization.

Prior work in multimodal domain generalization for action recognition includes RNA-Net \cite{planamente2022domain}, which balances audio and video feature norms, 
and SimMMDG \cite{dong2024simmmdg}, which employs contrastive learning and cross-modal translation to separate modality-specific and shared features. In contrast, our approach provides experimental analysis that highlights the resilience of audio and motion features to domain shifts, emphasizing their pivotal role in achieving more robust generalization compared to prior methods. CIR \cite{plizzari2023can} introduces ARGO1M for testing action recognition generalization across unseen scenarios and locations, using cross-instance reconstruction with text guidance, aligning visual narrations with appearance. Differently, we align audio with audio-based narrations, improving the robustness of representations. While \cite{oncescu2024sound} uses an LLM to generate audio-centric narrations from visual ones, we obtain audio narrations from audio with an audio captioner \cite{deshmukh2023pengi} and calculate consistency rate to determine how consistent audio is with video.

To summarize, our contributions are: (1) a multimodal framework that integrates motion, audio, and appearance features to advance domain generalization in first-person activity recognition, achieving state-of-the-art results on the ARGO1M dataset; (2) a thorough analysis highlighting the resilience of motion and audio features to domain shifts, emphasizing their pivotal role in domain generalization; (3) the alignment of audio narrations with audio features to strengthen the robustness of action representations; and (4) the use of consistency ratings between audio and visual narrations to optimize the influence of audio in prediction.

\section{Method}
\label{sec:method}

\subsection{Proposed Multimodal Setting}

Each training sample consists of a video clip paired with corresponding audio, visual narration and audio narration. The former is provided in the dataset we use \cite{plizzari2023can}, while we use Pengi \cite{deshmukh2023pengi} to generate the latter. 
While we adopt the approach from \cite{plizzari2023can} of using separate frozen encoders for each modality to extract base features, our method differs by incorporating the audio modality and utilizing distinct encoders for each modality, instead of a single encoder for fused appearance and motion features. Specifically, we train (from scratch) separate encoders $f_{appearance}$, $f_{motion}$, and $f_{audio}$ to derive domain-generalizable features $ap$, $m$, and $a$ for appearance, motion, and audio. Additionally, we train a separate encoder $f_{text}$ extracts the features of visual narration $t$ and audio narration $\hat{t}$. See Fig. 2.
The action prediction $\hat{y}$ is generated by the classifier $h$ based on the fused multimodal embedding.
The cross-entropy loss $\mathcal{L}_c$ is computed between the true and predicted action labels, $y$ and $\hat{y}$.

\subsection{Consistency Ratings to Enhance Fusion}


Before training, we calculate consistency ratings for each video sample using a LLM \cite{touvron2023llama}, assessing the degree to which the audio and visual narrations correspond semantically. 
This information is used to control fusion. 
Computing consistency between audio and visual content through text captures abstract concepts that raw features might miss. While features emphasize low-level details, text-based evaluations ensure that the audio and visual content correspond at a deeper, conceptual level. 


We use the following prompt obtain consistency ratings:

\advance\leftmargini -1em
\begin{quote}
\vspace{-0.2cm}
\ninept
Rate the consistency between two narrations from the same video out of 100. The first narration describes the visual aspect, and the second describes the audio. Consider how well the audio narration overlaps with and complements the visual narration. Output only the percentage score. 
\end{quote}
\vspace{-0.2cm}

The consistency ratings $r$ are then used as weights to modulate the contribution of the audio embeddings $a$ before the concatenation of appearance, motion, and audio features, as shown in Fig. 2. Specifically, only during training, the audio embeddings are scaled by the consistency rating $r$, such that $a = a * r$. This consistency-weighted audio approach ensures that audio information with strong semantic alignment to the visual content exerts a greater influence on the final prediction.
This approach enhances the quality of audio representations during training, minimizing the impact of noisy or irrelevant audio cues and improving the overall robustness.

\subsection{Text Guided Alignment}


Aligning visual features with 
narrations enriches the model with domain-invariant, human-like understanding, enhancing its ability to generalize across domains \cite{min2022grounding}. As discussed previously, 
we generate \emph{audio} narrations that are specifically tied to the audio content. 
The LLM in audio captioner Pengi \cite{deshmukh2023pengi} mimics human-like understanding, providing audio-specific, semantically rich descriptions. 
Since Pengi uses a separate encoder for audio feature extraction, its generated audio narrations are initially not aligned with our model's audio features $a$. 
Thus, in our approach, we align audio features $a$ with these audio narration features $\hat{t}$, while aligning appearance features $ap$ and motion features $m$ with visual narration features $t$ using contrastive learning.

Given a batch of samples $\mathcal{B} = \{(ap_i,m_i,a_i,t_i,\hat{t}_i)\}^{B}_{i=1}$, we frame the alignment as noise contrastive estimation \cite{oord2018representation}.
Specifically, $\mathcal{L}_{{ap} \rightarrow {t}}$ treats the appearance $ap_i$ as the anchor, with other narrative texts serving as negatives, and minimizes: 
\begin{equation}
    \mathcal{L}_{{ap} \rightarrow {t}} = -\dfrac{1}{|\mathcal{B}|} \sum^{B}_{i} log \dfrac{exp(s(ap_i,t_i)/ \tau)}{\sum^{B}_{j} exp(s(ap_j,t_j)/ \tau)}
\end{equation} 
where $s(\cdot, \cdot)$ is the cosine similarity and $\tau$ is a learnable temperature. In a similar manner, $\mathcal{L}_{{t} \rightarrow {ap}}$ uses the text $t_i$ as the anchor, with other appearance features as negatives. These losses are then combined to create the appearance-text alignment loss $\mathcal{L}_{ap,t} = \mathcal{L}_{{t} \rightarrow {ap}} + \mathcal{L}_{{ap} \rightarrow {t}}$. Likewise, motion-text alignment $\mathcal{L}_{m,t}$ and audio-text alignment $\mathcal{L}_{a,\hat{t}}$ are computed, and all these alignment losses are aggregated into a total alignment loss $\mathcal{L}_{align} = \mathcal{L}_{ap,t} + \mathcal{L}_{m,t} + \mathcal{L}_{a,\hat{t}}$.

To form the overall training objective, we combine the alignment loss with the cross-entropy classification loss:
\begin{equation}
\mathcal{L} = \mathcal{L}_{c} + \lambda \mathcal{L}_{align}
\end{equation}
where $\lambda = 0.1$ is used to weight the alignment loss.

\section{Experiments}
\label{sec:experiments}

\begin{table}[t]
\centering
\resizebox{\columnwidth}{!}{
\begin{tabular}{ccccccc}
\toprule
 Modality & Me & Co & Ar & Sh & Cl & Mean \\ 
 Setting & SAU & JPN & ITA & IND & US-MN &  \\ 
 
 \midrule

Audio* & 27.6 & 42.9 & 30.4 & 29.3 & 28.7 & 31.8  \\ 
Audio & 25.8 & 23.1 & 23.0 & 24.9 & 24.5 & 24.3  \\ 
 & (\textcolor{red}{-6.9\%}) & (\textcolor{red}{-85.7\%}) & (\textcolor{red}{-32.1\%}) & (\textcolor{red}{-21.6\%}) & (\textcolor{red}{-17.1\%}) & (\textcolor{red}{-32.7\%})  \\ 
 \midrule

Motion* & 27.0 & 29.3 & 32.0 & 29.6 & 26.2 & 28.9  \\ 
Motion & 25.9 & 19.1 & 21.9 & 26.8 & 22.8 & 23.3  \\ 
 & (\textcolor{red}{-4.2\%}) & (\textcolor{red}{-53.4\%}) & (\textcolor{red}{-46.1\%}) & (\textcolor{red}{-10.4\%}) & (\textcolor{red}{-14.9\%}) & (\textcolor{red}{-25.8\%})  \\ 
 \midrule

Appearance* & 34.9 & 56.4 & 51.2 & 43.0 & 39.2 & 44.9  \\ 
Appearance & 31.6 & 26.6 & 29.3 & 31.1 & 28.7 & 29.5  \\ 
 & (\textcolor{red}{-10.4\%}) & (\textcolor{red}{-112.0\%}) & (\textcolor{red}{-74.7\%}) & (\textcolor{red}{-38.2\%}) & (\textcolor{red}{-38.2\%}) & (\textcolor{red}{-54.8\%})  \\ 
 \midrule

Multimodal* & 36.5 & 59.0 & 52.5 & 44.9 & 40.4 & 46.7  \\ 
Multimodal & 34.7 & 30.7 & 31.8 & 34.2 & 32.4 & 32.7  \\
 & (\textcolor{red}{-5.1\%}) & (\textcolor{red}{-92.1\%}) & (\textcolor{red}{-65.1\%}) & (\textcolor{red}{-31.2\%}) & (\textcolor{red}{-24.7\%}) & (\textcolor{red}{-42.8\%})  \\ 

 \bottomrule
\end{tabular}}
\caption{Percentages show performance drop when the training set excludes samples from the test domain, vs when it includes them (*). Audio and motion exhibit less performance drop compared to appearance, highlighting their resilience to shifts.  
The baseline method (cross-entropy loss) is used.
}
\label{tab:table1}
\end{table}

\begin{table*}[t]
\centering
\begin{tabular}{ccccccccccccc}
\toprule
 Method  & Me & Co & Ar & Sh  & Cl & Pl & Sp & Ga & Bu & Kn & Mean \\ 
 & SAU & JPN & ITA & IND & US-MN &  US-IN & COL & US-PNA & US-PNA & IND &\\ 
 
 \midrule
 Baseline  &34.7& 30.7& 31.8& 34.2& 32.4& 36.1& 33.8& 34.6& 33.9& 28.9& 33.1 \\
 CIR \cite{plizzari2023can}  &35.6& \bf{32.4}& 32.4& 35.4& 33.1& 38.2& 34.7& 36.1& 35.4& 30.4& 34.3 \\
 Ours  &\bf{36.0}& 32.2& \bf{32.7}& \bf{36.0}& \bf{33.5}& \bf{38.3}& \bf{35.1}& \bf{36.5}& \bf{35.8}& \bf{30.6}& \bf{34.7} \\
 \bottomrule
\end{tabular}
\caption{A comparison with the state of the art on ARGO1M 
using appearance, motion, and audio.}
\label{tab:table2}
\end{table*}

\begin{table}[t]
\centering
\resizebox{\columnwidth}{!}{
\begin{tabular}{cccccccc}
\toprule
 Method & Modality & Me & Co & Ar & Sh & Cl & Mean \\ 
 & Setting & SAU & JPN & ITA & IND & US-MN &  \\ 
 \midrule
 Baseline & Ap & 31.6 & 26.6 & 29.3 & 31.1 & 28.7 & 29.5  \\ 
 Baseline & Ap-Mo & 32.7 &27.5 &30.0 &31.3 &29.5 & 30.2  \\ 
 Baseline & Ap, Mo & 33.2 & 27.8 & 30.8 & 31.7 & 29.7 & 30.6  \\ 
 Baseline & Ap, Mo, Au & \bf{34.7} &\bf{30.7}& \bf{31.8}& \bf{34.2}& \bf{32.4}& \bf{32.7}  \\ 
 
 \bottomrule
\end{tabular}}
\caption{Impact of combining different modalities—\textbf{Ap}pearance, \textbf{Mo}tion, and \textbf{Au}dio—
across domains. 
}
\label{tab:table3}
\end{table}

\begin{table}[t]
\centering
\resizebox{\columnwidth}{!}{
\begin{tabular}{ccccccc}
\toprule
 Method  & Me & Co & Ar & Sh & Cl & Mean \\ 
 & SAU & JPN & ITA & IND & US-MN &  \\ 
 \midrule
 Baseline ($B$) & 34.7 &30.7& 31.8& 34.2& 32.4& 32.7  \\ 
 $B +$ weighted $a$ by $r$ & 34.9& 30.9&31.9&34.4&32.4&32.9  \\ 
  $B +  \mathcal{L}_{vt} + \mathcal{L}_{mt} + \mathcal{L}_{at}$ & 35.4&31.6&32.3&35.2&32.9&33.5  \\ 
 $ B + \mathcal{L}_{vt} + \mathcal{L}_{mt} + \mathcal{L}_{a\hat{t}}$ & 35.8&31.9&32.6&35.6&33.3&33.8 \\ 
   
  Ours & \bf{36.0}&\bf{32.2}&\bf{32.7}&\bf{36.0}&\bf{33.5}&\bf{34.1}  \\ 
 
 \bottomrule
\end{tabular}}
\caption{Ablation 
showing the impact of alignment losses and consistency-weighted audio.  
All methods use (Ap, Mo, Au).}
\label{tab:table4}
\end{table}

\textbf{Implementation details.} We leverage the frozen pretrained SlowFast model \cite{feichtenhofer2019slowfast}, utilizing its slow pathway for capturing appearance features and its fast pathway for extracting motion features. For audio features, we employ BEATs \cite{chen2022beats}, while CLIP-ViT-B-32 \cite{radford2021learning} is used to extract text features. The trained encoders $f$ each consist of two fully connected layers, each followed by ReLU activation and Batch Normalization \cite{ioffe2015batch}. We use batch size 128 for 50 epochs with the Adam optimizer \cite{kingma2014adam}. The learning rate is initially set to 2e-4, with decay factor 10 applied at epochs 30 and 40.

\textbf{Dataset.} ARGO1M \cite{plizzari2023can} is an egocentric video dataset curated from Ego4D, specifically designed to analyze scenario and location-based domain shifts. The dataset includes 10 distinct train-test splits, ensuring that the domains of the samples—both scenario and location—do not overlap between the training and test sets. The test set scenarios and geographic locations include Gardening in Pennsylvania (Ga, US-PNA), Cleaning in Minnesota (Cl, US-MN), Knitting in India (Kn, IND), Shopping in India (Sh, IND), Building in Pennsylvania (Bu, US-PNA), Mechanic in Saudi Arabia (Me, SAU), Sport in Colombia (Sp, COL), Cooking in Japan (Co, JPN), Arts and Crafts in Italy (Ar, ITA), and Playing in Indiana (Pl, US-IN). ARGO1M contains 1,050,371 video clips. However, due to the absence of audio in some clips, we utilized approximately 60\% of the total dataset where audio is available. We report top-1 accuracy for each test split, along with the mean accuracy across the splits.

\subsection{Audio and Motion Resilience to Domain Shifts}

In Table~\ref{tab:table1}, the mean percentage drops in parentheses indicate the extent of domain shifts across various domains, comparing modality settings where the training set either contains samples that share location and/or scenario with the test domain, or share neither.
The mean performance drop for motion features is 25.8\%, while audio exhibits a drop of 32.7\%. In contrast, appearance shows a significantly higher shift (54.8\%).
This aligns with our hypothesis that temporal dynamics—such as consistent patterns of movement 
and continuity of sound—remain more stable across different environments and scenarios. In contrast, spatial semantics represented by appearance features, are more variable due to differences in objects, backgrounds, and other visual elements that can vary significantly from one domain to another. Moreover, the multimodal approach, which integrates audio, motion, and appearance, demonstrates a reduced shift with a mean drop of 42.8\% compared to appearance alone (54.8\%). Our analysis underscores the critical role of incorporating audio and motion features for robust domain generalization.

\subsection{Comparison with State of the Art}
Table~\ref{tab:table2} compares our proposed method, the state-of-the-art CIR \cite{plizzari2023can} approach, and a baseline model on the ARGO1M dataset. The baseline is trained solely with cross-entropy loss, whereas CIR enhances its performance through cross-instance reconstruction guided by text. Notably, although the original CIR results did not include audio, we have extended their approach by incorporating audio features and multiple trained encoders $f$ for each modality to ensure a fair comparison with our method. Our method consistently outperforms CIR across all domain splits, except for Co-JPN, achieving up to a 1.7\% improvement and a 1.2\% higher average performance overall. Additionally, our method outperforms the baseline by an average of 4.8\%.

\subsection{Ablations}
 Table~\ref{tab:table3} illustrates the impact of different modality settings on performance. In the `Ap-Mo' setting used by CIR \cite{plizzari2023can}, appearance and motion features are fused early and processed through a single encoder. We propose an alternative approach in the `Ap, Mo' setting, where separate encoders ($f_{appearance}$, $f_{motion}$) are employed to learn distinct domain-generalizable features, particularly because motion exhibits greater resistance to domain shifts. As a result, our `Ap, Mo' outperforms `Ap-Mo'. Furthermore, the integration of audio with separate encoders in `Ap, Mo, Au' yields the best overall performance.

Table ~\ref{tab:table4} evaluates the effectiveness of each component of our approach. Starting with the baseline (trained using only cross-entropy), we observe that weighting the audio embeddings $a$ by the consistency ratings $r$ enhances the model's performance, owing to the more reliable audio representations. Aligning all modalities with narration  (`$B +  \mathcal{L}_{vt} + \mathcal{L}_{mt} + \mathcal{L}_{at}$`) further improves performance.
Notably, when we align audio features with their corresponding audio narrations $\hat{t}$ using  $\mathcal{L}_{a\hat{t}}$ the model outperforms the version with $\mathcal{L}_{at}$ where audio features are aligned with the original visual-based narrations $t$. Finally, the combination of alignment losses and the consistency-weighted audio approach in our final method `Ours` achieves the best performance across all domains.

\textbf{Conclusion.}
We proposed a novel multimodal framework where motion, audio, and appearance improve domain generalization in first-person action recognition. We 
demonstrated that audio and motion features exhibit greater resilience to domain shifts than appearance features. By aligning audio features with audio-specific narrations and applying consistency-weighted audio during training, our method enhanced the robustness of action representations. 
\label{sec:conclusion}

\bibliographystyle{IEEEbib}
\bibliography{strings,refs}

\end{document}